\documentclass[10pt, a4paper]{article}
\usepackage{lrec}
\usepackage{multibib}
\newcites{languageresource}{Language Resources}

\usepackage{amssymb}
\usepackage{booktabs}
\usepackage{covington} 
\usepackage{enumitem}
\usepackage{graphicx}
\usepackage{epstopdf}
\usepackage[utf8]{inputenc}
\usepackage{latexsym}

\usepackage{subcaption}
\usepackage{tabularx}

\usepackage{soul}
\usepackage{url} 
\usepackage{xstring}

\newcommand{\EssentialArgument}{$\circledast$}
\newcommand{\SDExample}[2]{\begin{example}\small\textit{\textsf{#1}}\\[2pt]\textit{\textsf{#2}}\end{example}}
\newcommand{\SDInlineExample}[1]{{\small\textit{\textsf{``#1''}}}}
\title{A German Corpus for Fine-Grained Named Entity Recognition and Relation Extraction of Traffic and Industry Events}

\name{Martin Schiersch, Veselina Mironova, Maximilian Schmitt, Philippe Thomas, \\ {\bf \large Aleksandra Gabryszak, Leonhard Hennig}}

\address{DFKI GmbH \\
         Berlin, Germany \\
         \{firstname.lastname\}@dfki.de\\}

\abstract{
Monitoring mobility- and industry-relevant events is important in areas such as personal travel planning and supply chain management, but extracting events pertaining to specific companies, transit routes and locations from heterogeneous, high-volume text streams remains a significant challenge. This work describes a corpus of German-language documents which has been annotated with fine-grained geo-entities, such as streets, stops and routes, as well as standard named entity types. It has also been annotated with a set of 15 traffic- and industry-related n-ary relations and events, such as accidents, traffic jams, acquisitions, and strikes. The corpus consists of newswire texts, Twitter messages, and traffic reports from radio stations, police and railway companies. It allows for training and evaluating both named entity recognition algorithms that aim for fine-grained typing of geo-entities, as well as n-ary relation extraction systems.
\\ \newline \Keywords{Named Entity Recognition, Relation Extraction} }

\begin{document}

\maketitleabstract

\section{Introduction}
Monitoring relevant news and events is of central importance in many economic and personal decision processes, such as supply chain management~\cite{CHAE2015247}, market research~\cite{Mostafa2013}, and personal travel planning~\cite{schulz2013see}. Social media, news sites, and also more specialized information systems, such as online traffic and public transport information sources, provide valuable streams of text messages that can be used to improve decision making processes~\cite{hennig2016}. For example, a company's sourcing department may wish to monitor world-wide news for disruptive or risk-related events pertaining to their suppliers (e.g.\ natural disasters, strikes, liquidity risks), while a traveler wants to be informed about traffic events related to her itinerary (e.g.\ delays, cancellations). To fulfill such information needs, we need to extract events and relations from message streams that mention fine-grained entity types, such as companies, streets, or routes~\cite{yaghoobzadeh2017,shimaoka2017}. For example, from the sentence \SDInlineExample{Berlin: Rail replacement service between Schichauweg and Priesterweg on route S2}, we would like to extract a \emph{Rail Replacement Service} event with the arguments \emph{location}=\SDInlineExample{S2} of type \emph{location-route}, and \emph{start-loc}=\SDInlineExample{Schichauweg} respectively \emph{end-loc}=\SDInlineExample{Priesterweg} with types \emph{location-stop}.


Detecting such relations in textual message streams raises a number of challenges. Social media streams, such as Twitter, are written in a very informal, not always grammatically well-formed style~\cite{osborne2014}, which cannot easily be processed with standard linguistic tools. News sites provide well-formed texts, but their content is very heterogeneous and often hard to separate from non-relevant web page elements. 
Domain-specific information sources, like traffic reports, on the other hand, are topic-focused, but employ a wide variety of formats, from telegraph style texts to table entries. In addition, existing corpora for German-language Named Entity Recognition~\cite{tjong2003,benikova2014b} are mostly limited to standard entity types, and consist mainly of newswire and Wikipedia texts. These corpora also do not include annotations of events and relations.

In this work, we present a large German-language corpus consisting of documents from three different genres, namely newswire texts, Twitter, and traffic reports from radio stations, police and railway companies (Section~\ref{sec:collection}).
The documents have been annotated with fine-grained geo-entities, as well as standard entity types such as organizations and persons. In addition, the corpus has been annotated with a set of 15 mobility- and industry-related n-ary relation types (Section~\ref{sec:annotation}). Many of these relation and event types, such as accidents, traffic jams, and strike events, are not available in standard knowledge bases and hence cannot be learned in a distantly supervised fashion. The final corpus consists of $2,598$ documents with $22,075$ entity and $1,507$ relation annotations (Section~\ref{sec:corpus}). It allows for training and evaluating both named entity recognition algorithms that aim for fine-grained typing of geolocation entities, as well as for training of n-ary relation extraction systems.

\section{Dataset Collection}
\label{sec:collection}

 \begin{figure*}[ht!]
     \centering
     \begin{subfigure}[b]{0.3\textwidth}
         \includegraphics[width=\textwidth]{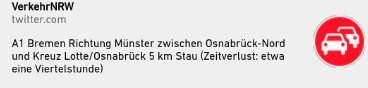}
         \caption{Example Twitter message}
     \end{subfigure}
     \begin{subfigure}[b]{0.3\textwidth}
         \includegraphics[width=\textwidth]{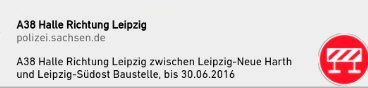}
         \caption{Excerpt of a RSS message}
     \end{subfigure}
     \begin{subfigure}[b]{0.3\textwidth}
         \includegraphics[width=\textwidth]{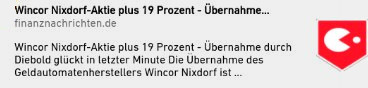}
         \caption{Excerpt of a news document}
     \end{subfigure}
     \caption{Examples for the three different data sources.}
     \label{fig:datasource-examples}
 \end{figure*}
\begin{table*}[ht!]
\small
\centering
\begin{tabular}{p{4.6cm}p{6.2cm}p{4.5cm}}
\toprule
 \textbf{Entity / Concept} & \textbf{Description} & \textbf{Examples} \\
\midrule
Location (\textsc{loc}) & General locations & Bayern, Zugspitze, Norden\\
Location-City (\textsc{loc-cit}) & Municipalities, e.g.\ cities, towns, villages & Berlin, Berlin-Buch, Hof\\
Location-Street (\textsc{loc-str}) & Named streets, highways, roads & Hauptstrasse, A1\\
Location-Route (\textsc{loc-rou}) & Named (public) transit routes & U1, ICE 557, Nürnberg -- Hof\\
Location-Stop (\textsc{loc-sto}) & Public transit stops, e.g.\ train stations, bus stops & S+U Pankow, Berlin-Buch\\
Organization (\textsc{org}) & General organizations & Greenpeace, Borussia Dortmund\\
Organization-Company (\textsc{org-com}) & The subset of organizations that are businesses & Siemens AG, BMW\\
Person  (\textsc{per}) & Persons & Angela Merkel\\
OrgPosition (\textsc{pos}) & A person's position within an organization & CEO, Vizepräsident\\
Date (\textsc{dat}) & Point in time, date & 1. September 2017, gestern\\
Time (\textsc{tim}) & Time of day & 8:30, 5 Uhr früh\\
Duration (\textsc{dur}) & Time periods & mehr als eine halbe Stunde\\
Distance (\textsc{dis}) & Distances with unit & 5 Kilometer\\
Number (\textsc{num}) & Other numeric entities, e.g.\ money, percentages & 3\%, 4 Millionen Euro \\
Disaster-Type (\textsc{dis-typ}) & Man-made and natural disaster types & Erdbeben, Überschwemmung \\
Trigger (\textsc{tri}) & Trigger terms or phrases for events & Stau, Streik, Entlassungen \\
\bottomrule
\end{tabular}
\caption{Definition of entity types annotated in the corpus.}
\label{tab:entities}
\end{table*}

To create the corpus, we collected a dataset of 3,789,803 tweets, 412,652 RSS feeds, and 860,307 news documents in the time period of Jan 1st, 2016 to March 31st, 2016. We aimed to collect only German-language texts by applying appropriate filter settings when crawling APIs, and by post-processing documents with langid.py~\cite{lui2012}. Figure~\ref{fig:datasource-examples} gives an example for each type of data source. All web documents, tweets, and RSS documents were transformed into a common Avro-encoded schema,\footnote{\url{avro.apache.org}} with fields for title, text, URI, and other attributes, as well as fields for the annotations. From this dataset, we randomly sampled documents from each data source for annotation.

\emph{Twitter}
We use the Twitter Search API\footnote{\url{dev.twitter.com/rest/public/search}} to obtain a topically focused streaming sample of tweets. We define the search filter using a list of approximately 150 mobility- and industry-relevant channels and 300 search terms. Channels include e.g.\ airline companies, traffic information sources, and railway companies. Search terms comprise event-related keywords such as ``traffic jam'' or ``roadworks'', but also major highway names, railway route identifiers, and airport codes. 

\emph{News}
We retrieve news pages and topically focused web sites using the uberMetrics Search API,\footnote{\url{doc.ubermetrics-technologies.com/api-reference/}} which provides an interface to more than 400 million web sources that are crawled on a regular basis. The API allows us to define complex boolean search queries to filter the set of web pages. We employ the same search terms as for Twitter, and limit the language to German. Boilerplate detection is used to remove extraneous contents from the HTML document~\cite{kohlschutter2010}. To speed up the annotation process, we limit each news document to the first 1000 characters, including the title, and discard the remainder of the text. Although this approach may result in the loss of some information, it is well known that in news writing, important information is presented first. The trimming may lead to incomplete final sentences, which annotators were advised to ignore.

\emph{RSS Feeds}
We implemented crawlers for a representative set of approximately 100 German-language RSS feeds that provide traffic and transportation information. Feed sources include federal and state police, radio stations, and air travel sources. The feeds were fetched at regular intervals during the 3-month period.

\section{Annotation Guidelines}
\label{sec:annotation}
For the targeted applications of supply chain monitoring and personal travel planning, we are interested in the annotation of fine-grained geo-locations, such as street names and public transport stops, as well as relation (or event) mentions with typically multiple arguments. That is, we are not only interested to know that a given event type occurred, such as a traffic jam, but also, on which road, between which exits, and what the resulting time delay is for drivers. We hence aim to recognize and extract n-ary ACE/ERE-style relations~\cite{doddington2004,ldc2015}. The  annotation guidelines and schema of the ACE Entities V6.6\footnote{\url{www.ldc.upenn.edu/sites/www.ldc.upenn.edu/files/english-entities-guidelines-v6.6.pdf}} and TimeML\footnote{\url{www.timeml.org/publications/timeMLdocs/annguide_1.2.1.pdf}} served as a basis for annotating standard entity types, such as organizations, persons and dates. 
The main difference to ACE guidelines is the treatment of geo-political entities (GPE) -- we chose to annotate them mainly as locations (LOC), and sometimes as organizations (ORG), in particular for cities, regions, or counties, as the relation types we are interested in typically refer to the location or organization aspects of a potential GPE entity. 

For the corpus annotation we use the markup tool Recon~\cite{Li2012}, which allows annotating n-ary relations among text elements. Recon provides a graphical user interface that enables users to mark arbitrary text spans as entities, to connect entities to create relations, and to assign semantic roles to argument entities. Each document was annotated by two trained annotators. In cases of disagreement, a third annotator was consulted to reach a final decision. We measured inter-annotator agreement for entity and relation annotations. For entity annotations, we evaluated agreement at the entity level by comparing labels and offsets. A high inter-annotator agreement thus implies that annotators agreed both on the extent of entities and their type. For relation mentions, we measured role and relation type agreement at the level of relation arguments for each annotated relation mention. Similar to entity inter-annotator agreement, arguments were identified based on the underlying concepts/entities and their character offsets. A high inter-annotator agreement hence means that annotators agreed on entity, entity extent, role, and relation type labels. Table~\ref{tab:anno-agreement} lists the inter-annotator agreement values of our corpus. The pairwise kappa agreement is moderate at around $0.58$ for entity annotations, which is somewhat lower than the $0.74$ reported by Benikova et al.~\shortcite{benikova2014b}. For relations, pairwise kappa agreement is $0.51$. 

\begin{table}[ht!]
\small
\centering
\begin{tabular}{lrr}
\toprule
 \textbf{Type} & \textbf{Cohen's} $\mathbf{\kappa}$ &  \textbf{Krippendorf's} $\mathbf{\alpha}$ \\
 \midrule
 Entities & 0.58 & 0.57 \\
 Relations & 0.51 & 0.45 \\
\bottomrule
\end{tabular}
\caption{Inter-annotator agreement for entities and relations.}
\label{tab:anno-agreement}
\end{table}

\subsection{Entities}
\label{sec:entities}

\begin{table*}[ht!]
\small
\centering
\begin{tabular}{p{3.5cm}p{12.8cm}}
\toprule
 \textbf{Relation / Event} & \textbf{Definition \& Arguments} \\
\midrule
\emph{Accident} &  Collision of a vehicle with another vehicle, person, or obstruction \newline \EssentialArgument location, \EssentialArgument trigger, delay, direction, start-loc, end-loc, start-date, end-date, cause (\textsc{tri}) \\\hline
\emph{Canceled Route} & Cancellation of public transport routes  \newline \EssentialArgument location (\textsc{loc-rou}), \EssentialArgument trigger, direction, start-loc, end-loc, start-date, end-date, cause (\textsc{tri}) \\\hline
\emph{Canceled Stop} & Cancellation of public transport stops \newline \EssentialArgument location (\textsc{loc-sto}), \EssentialArgument trigger, route, direction, start-date, end-date, cause (\textsc{tri}) \\\hline
\emph{Delay} & Delay resulting from remaining traffic disturbances \newline \EssentialArgument location, \EssentialArgument trigger, delay, direction, start-loc, end-loc, start-date, end-date, cause (\textsc{tri}) \\\hline
\emph{Disaster} & Sudden catastrophe causing great damage to structures or loss of life  \newline \EssentialArgument type, \EssentialArgument location, date, victims (\textsc{num}), damage-costs (\textsc{num}), trigger \\\hline
\emph{Obstruction} & Temporary installation to control traffic \newline \EssentialArgument location, \EssentialArgument trigger, delay, direction, start-loc, end-loc, start-date, end-date, cause (\textsc{tri}) \\\hline
\emph{Rail Replacement Service} & Replacement of a passenger train by buses or other substitute public transport services \newline \EssentialArgument location (\textsc{loc-rou}), \EssentialArgument trigger, delay, direction, start-loc, end-loc, start-date, end-date, cause (\textsc{tri}) \\\hline
\emph{Traffic Jam} & Line of stationary or very slow-moving traffic \newline \EssentialArgument location (\textsc{loc-str}), \EssentialArgument trigger, delay, jam-length, direction, start-loc, end-loc, start-date, end-date, cause (\textsc{tri}) \\\hline
\midrule
\emph{Acquisition} & Purchase of one company by another \newline \EssentialArgument buyer, \EssentialArgument acquired, seller, date, price, trigger \\\hline
\emph{Insolvency} & Insolvency of a company  \newline \EssentialArgument company, \EssentialArgument trigger, date, location \\\hline
\emph{Layoffs} & Layoffs from companies, including number of people fired.  \newline \EssentialArgument company, \EssentialArgument trigger, date, location, num-laid-off \\\hline
\emph{Merger} &  Merger of companies that is not a clear buy-up  \newline \EssentialArgument old (\textsc{org-com A}), old (\textsc{org-com B}), new (\textsc{org-com}), date, trigger \\\hline
\emph{Organization Leadership} & Relationship between an organization and its leaders, board members, directors, etc. \newline \EssentialArgument organization, \EssentialArgument person, position, from, to, trigger \\\hline
\emph{SpinOff} & Parent company ``splits off'' a section as a separate new company \newline  \EssentialArgument parent (\textsc{org-com}), \EssentialArgument child (\textsc{org-com}), location, trigger \\\hline
\emph{Strike} &  Strike action affecting a company or organization  \newline \EssentialArgument company, \EssentialArgument trigger, date, location, num-striking, striker, union (\textsc{org}) \\
\bottomrule
\end{tabular}
\caption{Definition of the 15 target relations of the domains \emph{Mobility} and \emph{Industry}. \EssentialArgument{} denotes the essential arguments of the relation that define the identity of a relation instance. Entity types are abbreviated or omitted in unambiguous cases.}
\label{tab:relations}
\end{table*}

Table~\ref{tab:entities} lists the entity types we currently annotate, and provides a brief explanation of each type. In general, annotators were advised to choose the more specific entity type for a given entity mention (e.g. organization-company instead of organization), unless it was unclear from the context whether the entity mention referred to the specific or the more general type. This is for example the case for traffic jam and accident reports on main highways, where the exits often use the name of the closest city, e.g.\ \SDInlineExample{accident on the A1 between [Bremen] and [Oldenburg]}. Here, \SDInlineExample{Bremen} and \SDInlineExample{Oldenburg} are ambiguous between the types \emph{location-city} and \emph{location-street}.\footnote{In the remainder of this document, `[' and `]' are used to denote the extent of an entity or relation mention}

\emph{Organization-Company} covers all commercial organizations, including media and entertainment businesses. It does not cover governmental or religious organizations, but may include sports teams. In general, though, we are interested in companies that provide services to other companies, e.g.\ in the form of products, parts, components, technologies, or non-physical services. 

\emph{Location} subtypes are of interest to pin-point exact locations for any event of interest, e.g.\ by a lookup in OpenStreetMap,\footnote{\url{openstreetmap.org}} and to distinguish between location types where necessary. We do not tag locations if they are used as metonyms for organizations or GPEs, as in the case of capital cities denoting the government of a country. 

In the case of traffic reports, we also consider directions, including cardinal points, as locations, for example:

\SDExample{Stau auf der B2 [stadteinwärts]}{(Traffic jam on the B2 [into town])}
\SDExample{Auf der A1 Nähe Münster Stau in [beiden Richtungen]}{(On the A1, near Münster, traffic jam in [both directions])}

For traffic-related locations, specifiers are included in the mention extent when required, e.g. \SDInlineExample{[Kreuz München-Nord]} (\SDInlineExample{[Cross Munich-North]}) as well as \SDInlineExample{[Dreieck Havelland]} (\SDInlineExample{[Junction Havelland]}), \SDInlineExample{[Anschlussstelle Adlershof]} (\SDInlineExample{[Exit Adlershof]}), \SDInlineExample{[Abzweig nach Basel]} (\SDInlineExample{[Branch to Basel]}), etc. Similarly, terms like \SDInlineExample{Kreis} (\SDInlineExample{county}) in \SDInlineExample{[Kreis Tuttlingen]} are included to distinguish the county from the city. However, terms like \SDInlineExample{Ecke} (\SDInlineExample{corner}) or \SDInlineExample{Kreuzung} (\SDInlineExample{intersection}) are not included in the extent of \emph{Location-Street} entities, because they are not an integral part of the location's name.

City names that occur in transit routes are labeled as \emph{Location-City} when they are used to indicate the direction of the route, and as \emph{Location-Stop}s in every other case. In the case of highway exits, city names are labeled as \emph{Location}s, since they actually denote the exit (and its geographic position), and not the city. For flight routes, we chose to label city names as \emph{Location-City} unless the reference includes the specific airport used, e.g.\ \SDInlineExample{Heathrow} or \SDInlineExample{MUC}.

\emph{Location-Route}s are either generic transit lines (e.g. \SDInlineExample{S2}), or a specific instance of this line (\SDInlineExample{the next S2 which was supposed to arrive at 19:40}). In general, they are referred to by letter-number combinations, but sometimes consist of concatenated stop or city names:

\SDExample{[Günzburg -- Mindelheim]: Störung an einem Bahnübergang}{ ([Günzburg - Mindelheim]: Disruption at a crossing)}
\SDExample{Ersatzverkehr auf der Linie [RE 3] [Stralsund/Schwedt (Oder) - Berlin – Elsterwerda]}{(Rail replacement service on the route [RE 3] [Stralsund/Schwedt (Oder) - Berlin - Elsterwerda])}

Common nouns and noun phrases are annotated like proper names as entities of the corresponding type. Most often, they are used to denote a group of entities, e.g.:

\SDExample{Die [EC-Züge] zwischen Dresden Hbf und Praha hl.n. fallen aus}{(The [EC trains] between Dresden main station and Prague main station are cancelled)}
\SDExample{Fraport übernimmt [14 griechische Flughäfen]}{(Fraport acquires [14 Greek airports])}

\emph{Trigger} concepts are a generic class of annotations that cover terms or phrases that indicate a specific event type, and that sometimes are required to create at least a binary relation mention within a sentence. For example, given the message \SDInlineExample{Stau auf der Warschauer Strasse} (\SDInlineExample{Traffic jam on Warschauer street}), the location \SDInlineExample{Warschauer Strasse} alone is not sufficient to annotate a \emph{Traffic Jam} event, which requires the additional annotation of the trigger \SDInlineExample{Stau} to distinguish it from other traffic-related event types. This reasoning applies for the relations \emph{Insolvency}, \emph{Layoffs} and \emph{Strike} of the \emph{Industry} domain, and for all relations of the \emph{Mobility} domain except for the relation \emph{Disaster}. However, the argument \emph{type} of the relation \emph{Disaster} can be filled only with concepts which can be considered as triggers for this event (earthquake, flood, nuclear accidents, etc.). The majority of mobility-related events we are interested in follow this pattern of \emph{Location} $+$ \emph{Trigger} (or in the case of industry-related events, \emph{Company} $+$ \emph{Trigger}) to distinguish between different event types that are expressed using the same syntactic patterns (see Section~\ref{sec:events}).

Punctuation characters, such as ``-'', ``/'', ``\#'' and ``@'' are not included in the mention extent unless they occurred inside a multi-token entity, e.g.\ \SDInlineExample{\#[Flughafen \#Tempelhof]}. Annotators were advised to make the mention extent as long as required to accurately denote a specific entity. The extent could include adjectives, numerals (\SDInlineExample{more than}, \SDInlineExample{a few}, \SDInlineExample{several}), or numbers, if these were used to denote a specific subset of a set-based named entity mention. 

If an entity was referred to by two or more token sequences, e.g.\ \SDInlineExample{Volkswagen (VW)}, \SDInlineExample{A1 Bremen - Hamburg}, the annotators were advised to annotate two separate entities as in \SDInlineExample{[Volkswagen] ([VW])}. 

As a rule, unless required for annotating a relation mention, nested entity mentions were not annotated, e.g.\ in \SDInlineExample{PD Zwickau} (\SDInlineExample{police department Zwickau}), \SDInlineExample{[PD Zwickau]} was labeled as an \emph{Organization}, but the nested \SDInlineExample{Zwickau} was not labeled as a \emph{Location-City}. 

\subsection{Relations and Events}
\label{sec:events}
We annotated two different sets of relations and events in the corpus, based on the requirements of the project this corpus was developed in. The first group of relations are mobility-related, and include for example \emph{Traffic Jam}s, \emph{Accident}s and \emph{Disaster}s. The second group of relations concerns companies, and includes e.g.\ \emph{Acquisition}, \emph{Strike} and \emph{Insolvency} events. Table~\ref{tab:relations} lists all relation types, together with their definitions and arguments. All relations have a set of required (typically two) and a set of optional arguments. For example, the relation \emph{Acquisition} has required arguments \emph{buyer} and \emph{acquired}, and optional arguments \emph{date}, \emph{price}, and \emph{seller}. The following examples illustrate our n-ary relation annotations:\\

Relation examples of the \textit{Mobility} domain

\SDExample{\emph{Accident}: [A8]\textsubscript{loc} Augsburg Richtung [München]\textsubscript{dir} - Schwerer [Unfall]\textsubscript{tri} - kurz vor [Ausfahrt Dasing]\textsubscript{sta}.}{(A8 from Augsburg to Munich - a severe accident - just before the exit Dasing)}

\SDExample{\emph{Canceled Route}: Wegen des Warnstreiks hat die Lufthansa [mehrere Flüge]\textsubscript{loc} in [Hamburg]\textsubscript{sta}, [Hannover]\textsubscript{sta} und weiteren Flughäfen [gestrichen]\textsubscript{tri}.}{(Because of the warning strike, Lufthansa has canceled several flights in Hamburg, Hanover and other airports)}

\SDExample{\emph{Canceled Stop}: Rinjani macht Ärger: [Flughafen auf Bali]\textsubscript{loc} wegen [Vulkanausbruch]\textsubscript{cau} [gesperrt]\textsubscript{tri}.}{(Rinjani causes trouble: Bali airport is closed due to volcanic eruption.)}

\SDExample{\emph{Delay}: [S-Bahn-Verkehr Stuttgart]\textsubscript{loc}: [Notarzteinsatz]\textsubscript{cau} in [Feuerbach]\textsubscript{sta} sorgt für [Verspätungen]\textsubscript{tri}}{(S-Bahn traffic Stuttgart: Emergency medical service in Feuerbach causes delays)}

\SDExample{\emph{Disaster}: [Mehrere Tote]\textsubscript{vic} bei erneutem [Erdbeben]\textsubscript{typ} in [Japan]\textsubscript{loc}}{(Several dead in another earthquake in Japan)}

\SDExample{\emph{Obstruction}: Wegen [Notarzteinsatzes]\textsubscript{cau} ist derzeit die [Strecke]\textsubscript{loc} zwischen [Gerlenhofen]\textsubscript{sta} und [Senden]\textsubscript{end} [gesperrt]\textsubscript{tri}. }{(Due to an emergency medical service, the route between Gerlenhofen and Senden is currently closed)}

\SDExample{\emph{Rail Replacement Service:} [RB59]\textsubscript{loc}: Vom [11.6.]\textsubscript{sdat} - [3.7.]\textsubscript{edat} [Schienenersatzverkehr]\textsubscript{tri} zwischen [Soest]\textsubscript{sta} und [Holzwickede]\textsubscript{end} im Spätverkehr.}{(RB59: Rail replacement service from 6/11 until 7/3 between Soest and Holzwickede during evening hours.)}

\SDExample{\emph{Traffic Jam}: [A40]\textsubscript{loc} Duisburg Richtung [Venlo]\textsubscript{dir} zwischen [Neukirchen- Vluyn]\textsubscript{sta} und [Kempen]\textsubscript{end} [10 km]\textsubscript{len} [Stau]\textsubscript{tri}}{(A40 Duisburg - Dortmund between Neukirchen- Vluyn and Kempen 10 km traffic jam)}

Relation examples of the \textit{Industry} domain

\SDExample{\emph{Acquisition}: Wirecard AG und ihre Tochtergesellschaft [Wirecard Acquiring \& Issuing]\textsubscript{buy} haben den Zahlungsdienstleister [Moip Pagamentos]\textsubscript{acq} [übernommen]\textsubscript{tri}.}{(Wirecard AG and its subsidiary Wirecard Acquiring \& Issuing have acquired the payment service provider Moip Pagamentos.)}

\SDExample{\emph{Insolvency}: [Imtech]\textsubscript{com}  [Insolvenz]\textsubscript{tri}  gefährdet BER-Eröffnung}{(Imtech insolvency endangers BER opening)}

\SDExample{\emph{Layoffs}: [Entlassungen]\textsubscript{tri} bei [Credit Agricole Indosuez]\textsubscript{com} in [Genf]\textsubscript{loca}}{(Layoffs at Credit Agricole Indosuez in Genf)}

\SDExample{\emph{Merger}: Der Panzerhersteller [Krauss-Maffei Wegmann]\textsubscript{old} besiegelt den [Zusammenschluss]\textsubscript{tri} mit dem französischen Rüstungskonzern [Nexter]\textsubscript{old}.}{(Tank manufacturer Krauss-Maffei Wegmann seals merger with French arms company Nexter.)}

\SDExample{\emph{Organization Leadership}: [Bernd Hansen]\textsubscript{per}, CEO\textsubscript{pos} [Hansen Gruppe]\textsubscript{com}}{(Bernd Hansen, CEO Hansen Group)}

\SDExample{\emph{SpinOff}: [Kölnische Unfall-Versicherungs- Aktiengesellschaft zu \#Köln a.Rhein]\textsubscript{chi}, gegr. [1919]\textsubscript{dat} als [Ableger]\textsubscript{tri} der [Colonia]\textsubscript{par}}{(Kölnische Unfall-Versicherungs-Aktiengesellschaft zu \#Köln a.Rhein, est. in 1919 as a spin-off of the Colonia)}

\SDExample{\emph{Strike}: Am [Freitag]\textsubscript{dat} haben die [Amazon]\textsubscript{com}- Mitarbeiter im [Leipziger]\textsubscript{loc} Versandzentrum des Unternehmens erneut [gestreikt]\textsubscript{tri}.}{(On Friday, Amazon employees in the company's shipping center in Leipzig once more went on strike.)}

Some of the relations are semantically related to each other and can occur together even in very short texts such as tweets or RSS feeds. For example, the relation \emph{Traffic Jam} often correlates with \emph{Accident} and \emph{Obstruction} relation mentions. This also applies to the relation \emph{Obstruction} and the event \emph{Disaster}, and \emph{Delay} relations and the events \emph{Canceled Route}, \emph{Canceled Stop} and \emph{Rail Replacement Service}. In the \emph{Industry} domain, we observe that reports of corporate events often include information about leaders of an organization, i.e.\ a \emph{Organization Leadership} relation is mentioned together with another relation. 

The annotators annotated only explicitly expressed relation mentions where all arguments -- required and optional -- occurred within a single sentence. In cases of multiple occurrences of an argument, they chose the arguments occurring within the shortest overall text span. Future or planned relations, such as potential acquisitions or announced strikes, were also marked up, and labeled with an additional attribute to indicate this status. Negated relation mentions (e.g.\ a canceled acquisition), or events marking the end of a relation (e.g.\ \SDInlineExample{Traffic jam has dissolved}) were not annotated.\footnote{However, the files containing such mentions were marked by renaming them. Fully annotating the negated relation mentions and including them in the corpus remains future work.} The following examples illustrate the three types of relation mentions:

\SDExample{\emph{Factual}: Der Panzerhersteller Krauss-Maffei Wegmann besiegelt den Zusammenschluss mit dem französischen Rüstungskonzern Nexter.}{(Tank manufacturer Krauss-Maffei Wegman seals merger with French armaments group Nexter.)}
\SDExample{\emph{Potential}: Größer als BASF: US-amerikanische Chemie-Unternehmen DuPont und Dow Chemical planen Mega-Fusion.}{(Larger than BASF: US-American  chemical companies DuPont and Dow Chemical are planning mega-merger)}
\SDExample{\emph{Negation}: BHF-Bank: Fosun beteuert, keine Fusion mit Hauck \& Aufhäuser anzustreben}{(BHF-Bank: Fosun re-affirms not seeking a merger with Hauck \& Aufhäuser)}

\section{Corpus Statistics}
\label{sec:corpus}



\begin{table}[t!]
\centering
\small
\begin{tabular}{lrrrr}
\toprule
 & \textbf{News} & \textbf{Twitter} & \textbf{RSS} & \textbf{Total} \\
\midrule
Documents & 835 & 1,138 & 625 & 2,598\\
Sentences & 5,951 & 1,842 & 1,031 & 8,824\\
Sentences (avg.) & 7.13 & 1.62 & 1.65 & 3.40\\
Words & 113,089 & 19,558 & 19,595 & 152,242\\
Words (avg.) & 135.44 & 17.19 & 31.35 & 58.60\\
\bottomrule
\end{tabular}
\caption{Corpus Statistics}
\label{tab:dataset}
\end{table}

\begin{table}[t!]
\centering
\begin{tabular}{lrrrr}
\toprule
 & \textbf{News} & \textbf{Twitter} & \textbf{RSS} & \textbf{Total} \\
\midrule
Entities & 13,500 & 3,478 & 5,097 & 22,075\\
Entities (avg.) & 16.17 & 3.06 & 8.16 & 8.50\\
Relations & 597 & 454 & 456 & 1,507\\
Relations (avg.) & 0.71 & 0.40 & 0.73 & 0.58\\
\bottomrule
\end{tabular}
\caption{Annotation Statistics}
\label{tab:annotations}
\end{table}

 \begin{figure*}[t!]
  \centering
  \begin{subfigure}[b]{0.45\textwidth}
   \includegraphics[width=\columnwidth,clip,trim=0 0 0 0]{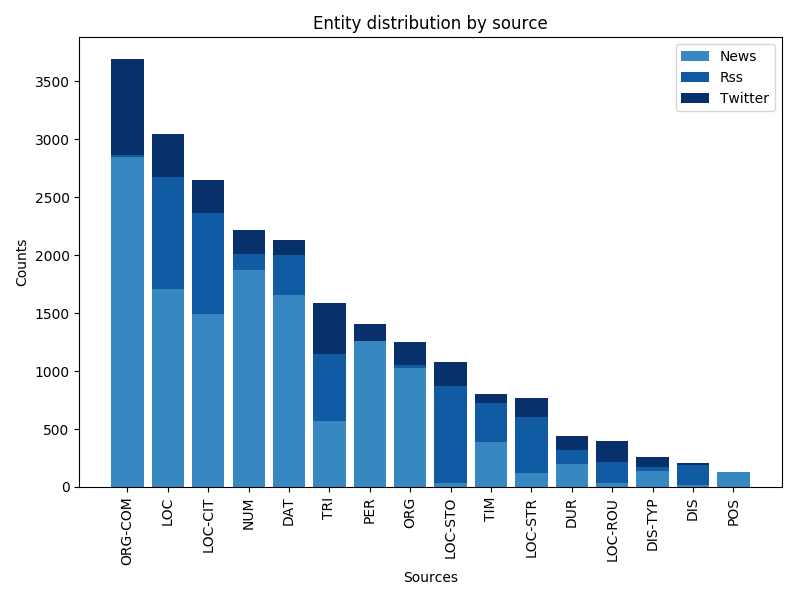}%
   \caption{Distribution of annotated entities}
   \label{fig:entity_dist}
  \end{subfigure}
 \begin{subfigure}[b]{0.45\textwidth}
   \includegraphics[width=\columnwidth,clip,trim=0 0 0 0]{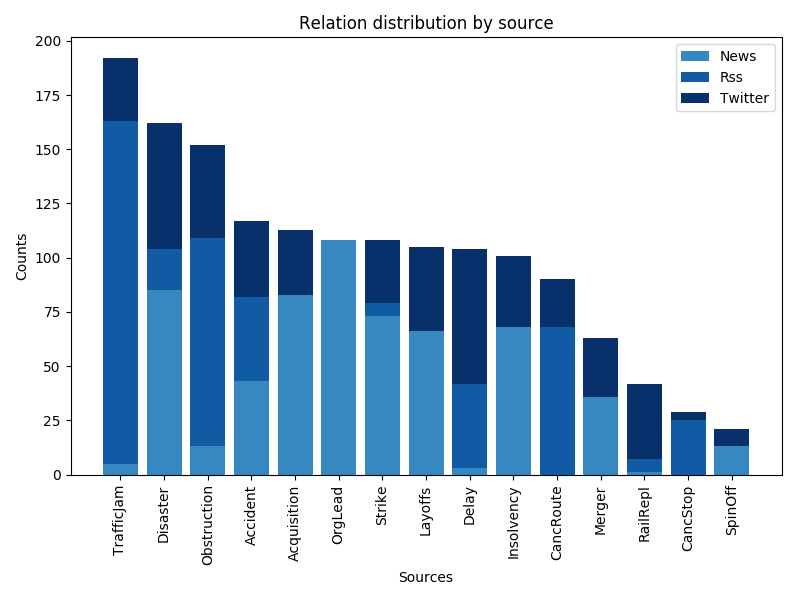}%
   \caption{Distribution of annotated relations}
   \label{fig:relation_dist}
   \end{subfigure}
   \caption{Entity and relation type distribution across document source types}
   \label{fig:annotation_dist}
 \end{figure*}

This section summarizes the key characteristics of the final corpus. It contains a total of $2,598$ documents with more than $150,000$ words. Table~\ref{tab:dataset} shows a brief summary of all document types, while Table~\ref{tab:annotations} summarizes the annotation statistics per document type. In total, the annotators labeled $22,075$ entities and $1,507$ relation occurrences. Due to their greater length, news documents contain the largest number of entity mentions, significantly more than the other two document types. Twitter documents on average contain fewer relation mentions than RSS and news documents. The overall fraction of documents containing at least a single relation mention is $58\%$, a rather high figure that can be attributed to the focused retrieval process which was used to create the initial dataset. 

Figure~\ref{fig:annotation_dist} shows the distribution of annotated entities and relations across document types. Companies, general locations, and cities are the most frequent entity types in our dataset. Public transit stops, streets, and routes are less frequently mentioned, and occur predominantly in tweets and RSS traffic reports. This is an expected distribution, since major news outlets generally do not report on day-to-day, local traffic events. With regards to relation types, traffic events like \emph{Traffic Jam}s and \emph{Obstructions} occur very frequently. Other event types occur with lower frequency in our annotated data, in particular, the annotators identified only very few instances of \emph{Canceled Stop} and \emph{SpinOff} events.

\subsection{Baseline NER and RE experiments}
We conducted a series of experiments to report initial performance figures on the presented corpus for the tasks of named entity recognition and relation extraction. We use the Stanford  CoreNLP tools~\cite{manning2014} for training a NER classifier, and a dependency pattern based model for relation extraction. 
The relation extraction algorithm, DARE, learns minimal dependency subgraphs that connect all relation arguments, and is described in~\cite{xu2007b,krause2012}. We did not perform any filtering of the extracted dependency patterns, i.e.\ we include all learned patterns, even ambiguous or low-frequency ones, in the model.\footnote{Obviously, properly filtering patterns may significantly improve performance, but state-of-the-art RE performance is not the goal of this study.} 

We randomly split the full dataset into $50\%$ training and $50\%$ test data. The NER model was trained using the standard feature configuration employed by Stanford CoreNLP for NER.\footnote{See \url{nlp.stanford.edu/software/crf-faq.html}} The RE models were trained with and without using gold standard NE annotations. 

For NER, we evaluated the model's performance both at the token level and at the concept level. The results are shown in Table~\ref{tab:results-ner}. We see that the overall token-level F1 score is close to $0.85$, a respectable figure given the confusability of location subtypes such as streets, stops, and routes in our dataset. For \emph{organization} and \emph{organization-company} entities, the average token-level F1 score is lower at approximately $0.78$, but for \emph{location} and its subtypes, it lies between $0.85 - 0.92$ (not shown).

\begin{table}[ht!]
\centering
\begin{tabular}{lrrr}
\toprule
  \textbf{Evaluation Type} & \textbf{Precision} & \textbf{Recall} & \textbf{F1} \\ 
\midrule
CRF (token) & $0.8966$ & $0.8024$ & $0.8469$ \\
CRF (concept) & $0.7984$ & $0.6797$ & $0.7343$ \\
\bottomrule
\end{tabular}
\caption{Performance of a standard CRF-based NER classifier on the presented dataset.}
\label{tab:results-ner}
\end{table}

For relation extraction, the models were evaluated at the mention level, by comparing predicted relation mentions with gold relation mentions. Since our dataset contains n-ary relations with optional and required arguments, we chose a soft matching strategy that counts a predicted relation mention as correct if all predicted arguments also occur in the corresponding gold relation mention, and if all required arguments have been correctly predicted, based on their role, underlying entity, and character offsets / extent. Optional arguments from the gold relation mention that are not contained in the predicted relation mention do not count as errors. In other words, we count a predicted relation mention as correct if it contains all required arguments and is subsumed by or equal to the gold relation mention.

Table~\ref{tab:results-re} shows the results of two RE evaluation runs, once with gold-standard NE annotations, and once without any gold annotations. As can be expected, the performance of the RE models using gold-standard NE annotations is significantly higher than that of the models using the trained NER classifier. The dependency-based DARE model achieves an F1 score of $0.28$ using gold-standard NEs, and is biased toward high-precision patterns, at the expense of recall.

\begin{table}[ht!]
\centering
\begin{tabular}{lrrr}
\toprule
  \textbf{Model} & \textbf{Precision} & \textbf{Recall} & \textbf{F1} \\ 
\midrule
DARE (CRF NE) & $0.4670$ & $0.1308$ & $0.2043$ \\
DARE (Gold NE) & $0.5274$	& $0.1923$	& $0.2818$ \\
\bottomrule
\end{tabular}
\caption{Performance of a dependency pattern based RE model on the presented dataset.}
\label{tab:results-re}
\end{table}

\section{Related Work}
There are very few available NER and RE datasets for German. Most noteworthy are the NER dataset presented by Benikova et al.~\shortcite{benikova2014b}, and the CoNLL-2003 dataset~\cite{tjong2003}. Both datasets contain annotations only for the three standard entity types, PER, ORG and LOC, as well as MISC/OTHER. The dataset by Benikova et al.\ includes nested annotations, whereas in our corpus, nested annotations are only annotated if they are required for a relation mention. Both datasets use news articles as their data source, with Benikova et al.'s dataset including Wikipedia texts. In contrast, our dataset also contains Twitter texts, as well as telegraphese-style reports from official traffic channels, which allows for text genre-specific evaluation of NER approaches. 

The ACE datasets~\cite{doddington2004,ldc2015} are similar to the dataset presented in this paper in that they include both NE and RE annotations. The various ACE datasets developed over the years consider a wide range of entity types, such as PER, ORG, LOC, GPE and FAC. Similarly, a range of different relation types are annotated in these datasets, including geographical, social and business relationships. However, all relations definitions are limited to binary relations, whereas our corpus contains n-ary relation mentions. None of the ACE datasets cover German-language documents.

Other well-known English relation extraction datasets include the corpora prepared for the TAC-KBP challenges~\cite{ji2011,surdeanu2013}, the SemEval-2010 Task 8 dataset~\cite{hendrickx2010}, and the TACRED dataset by~\cite{zhang2017}. 

\section{Conclusion}
We presented a corpus of German Twitter, news and traffic report texts that has been annotated with fine-grained geo-entities as well as a set of mobility- and industry-related events. Many of the event types annotated in the corpus are not available in standard knowledge bases, such as accidents, traffic jams, and strike events. We make the corpus and the guidelines available to the community at \url{https://dfki-lt-re-group.bitbucket.io/smartdata-corpus}. The dataset is distributed in an AVRO-based compact binary format, along with the corresponding schema and reader tools.

\section*{Acknowledgments}
This  research  was  partially  supported  by  the  German  Federal  Ministry  of  Economics  and Energy   (BMWi)   through   the   projects  SD4M (01MD15007B) and SDW (01MD15010A), by the German Federal Ministry of Transport and Digital Infrastructure through the project DAYSTREAM (19F2031A), and by the German Federal Ministry of Education and Research (BMBF) through the project BBDC (01IS14013E).

\section{Bibliographical References}
\label{main:ref}

\bibliographystyle{lrec}
\bibliography{article}

\begin{thebibliography}{}

\bibitem[\protect\citename{Benikova \bgroup et al.\egroup }2014]{benikova2014b}
Benikova, D., Biemann, C., and Reznicek, M.
\newblock (2014).
\newblock {NoSta-D} {Named} {Entity} {Annotation} for {German}: {Guidelines}
  and {Dataset}.
\newblock In Nicoletta Calzolari, et~al., editors, {\em Proceedings of the
  Ninth International Conference on Language Resources and Evaluation
  (LREC'14)}, pages 2524--2531, Reykjavik, Iceland, May. European Language
  Resources Association (ELRA).
\newblock ACL Anthology Identifier: L14-1251.

\bibitem[\protect\citename{Chae}2015]{CHAE2015247}
Chae, B.~K.
\newblock (2015).
\newblock {Insights from hashtag \#supplychain and Twitter Analytics:
  Considering Twitter and Twitter data for supply chain practice and research}.
\newblock {\em International Journal of Production Economics}, 165(Supplement
  C):247 -- 259.

\bibitem[\protect\citename{Doddington \bgroup et al.\egroup
  }2004]{doddington2004}
Doddington, G., Mitchell, A., Przybocki, M., Ramshaw, L., Strassel, S., and
  Weischedel, R.
\newblock (2004).
\newblock The {Automatic} {Content} {Extraction} ({ACE}) {Program} - {Tasks},
  {Data}, and {Evaluation}.
\newblock In {\em Proc. of {LREC}}.

\bibitem[\protect\citename{Hendrickx \bgroup et al.\egroup
  }2010]{hendrickx2010}
Hendrickx, I., Kim, S.~N., Kozareva, Z., Nakov, P., Ó~Séaghdha, D., Padó,
  S., Pennacchiotti, M., Romano, L., and Szpakowicz, S.
\newblock (2010).
\newblock {SemEval}-2010 {Task} 8: {Multi}-{Way} {Classification} of {Semantic}
  {Relations} between {Pairs} of {Nominals}.
\newblock In {\em Proceedings of the 5th {International} {Workshop} on
  {Semantic} {Evaluation}}, pages 33--38, Uppsala, Sweden, July. Association
  for Computational Linguistics.

\bibitem[\protect\citename{Hennig \bgroup et al.\egroup }2016]{hennig2016}
Hennig, L., Thomas, P., Ai, R., Kirschnick, J., Wang, H., Pannier, J.,
  Zimmermann, N., Schmeier, S., Xu, F., Ostwald, J., and Uszkoreit, H.
\newblock (2016).
\newblock Real-time discovery and geospatial visualization of mobility and
  industry events from large-scale, heterogeneous data streams.
\newblock In {\em Proceedings of ACL-2016 System Demonstrations}, pages 37--42,
  Berlin, Germany, August. Association for Computational Linguistics.

\bibitem[\protect\citename{Ji \bgroup et al.\egroup }2011]{ji2011}
Ji, H., Grishman, R., and Dang, H.~T.
\newblock (2011).
\newblock Overview of the {TAC} 2011 {Knowledge} {Base} {Population} {Track}.
\newblock In {\em Proc. of the 4th {Text} {Analysis} {Conference}}.

\bibitem[\protect\citename{Kohlsch\"{u}tter \bgroup et al.\egroup
  }2010]{kohlschutter2010}
Kohlsch\"{u}tter, C., Fankhauser, P., Nejdl, W., Kohlsch\"{u}tter, C.,
  Fankhauser, P., and Nejdl, W.
\newblock (2010).
\newblock {Boilerplate Detection Using Shallow Text Features}.
\newblock In {\em Proc. of WSDM}, pages 441--450.

\bibitem[\protect\citename{Krause \bgroup et al.\egroup }2012]{krause2012}
Krause, S., Li, H., Uszkoreit, H., and Xu, F.
\newblock (2012).
\newblock {Large-Scale} {Learning} of {Relation-Extraction} {Rules} with
  {Distant} {Supervision} from the {Web}.
\newblock In {\em Proc. of {ISWC}}, pages 263--278.

\bibitem[\protect\citename{Li \bgroup et al.\egroup }2012]{Li2012}
Li, H., Cheng, X., Adson, K., Kirshboim, T., and Xu, F.
\newblock (2012).
\newblock Annotating opinions in german political news.
\newblock In {\em 8th ELRA Conference on Language Resources and Evaluation}.
  European Language Resources Association (ELRA), 5.

\bibitem[\protect\citename{{Linguistic Data Consortium}}2015]{ldc2015}
{Linguistic Data Consortium}.
\newblock (2015).
\newblock Rich {ERE} annotation guidelines overview.
\newblock
  \url{http://cairo.lti.cs.cmu.edu/kbp/2015/event/summary_rich_ere_v4.1.pdf}.

\bibitem[\protect\citename{Lui and Baldwin}2012]{lui2012}
Lui, M. and Baldwin, T.
\newblock (2012).
\newblock langid.py: An off-the-shelf language identification tool.
\newblock In {\em Proc. of ACL: System Demonstrations}, pages 25--30.

\bibitem[\protect\citename{Manning \bgroup et al.\egroup }2014]{manning2014}
Manning, C., Surdeanu, M., Bauer, J., Finkel, J., Bethard, S., and McClosky, D.
\newblock (2014).
\newblock The {Stanford} {CoreNLP} {Natural} {Language} {Processing} {Toolkit}.
\newblock In {\em Proceedings of 52nd {Annual} {Meeting} of the {Association}
  for {Computational} {Linguistics}: {System} {Demonstrations}}, pages 55--60,
  Baltimore, Maryland, June. Association for Computational Linguistics.

\bibitem[\protect\citename{Mostafa}2013]{Mostafa2013}
Mostafa, M.~M.
\newblock (2013).
\newblock More than words: Social networks’ text mining for consumer brand
  sentiments.
\newblock {\em Expert Systems with Applications}, 40(10):4241 -- 4251.

\bibitem[\protect\citename{Osborne \bgroup et al.\egroup }2014]{osborne2014}
Osborne, M., Moran, S., McCreadie, R., Von~Lunen, A., Sykora, M., Cano, E.,
  Ireson, N., Macdonald, C., Ounis, I., He, Y., Jackson, T., Ciravegna, F., and
  O'Brien, A.
\newblock (2014).
\newblock Real-time detection, tracking, and monitoring of automatically
  discovered events in social media.
\newblock In {\em Proc. of ACL: System Demonstrations}, pages 37--42.

\bibitem[\protect\citename{Schulz \bgroup et al.\egroup }2013]{schulz2013see}
Schulz, A., Ristoski, P., and Paulheim, H.
\newblock (2013).
\newblock I see a car crash: Real-time detection of small scale incidents in
  microblogs.
\newblock In {\em Extended Semantic Web Conference}, pages 22--33. Springer,
  Berlin, Heidelberg.

\bibitem[\protect\citename{Shimaoka \bgroup et al.\egroup }2017]{shimaoka2017}
Shimaoka, S., Stenetorp, P., Inui, K., and Riedel, S.
\newblock (2017).
\newblock Neural {Architectures} for {Fine}-grained {Entity} {Type}
  {Classification}.
\newblock In {\em Proceedings of the 15th {Conference} of the {European}
  {Chapter} of the {Association} for {Computational} {Linguistics}: {Volume} 1,
  {Long} {Papers}}, pages 1271--1280, Valencia, Spain, April. Association for
  Computational Linguistics.

\bibitem[\protect\citename{Surdeanu}2013]{surdeanu2013}
Surdeanu, M.
\newblock (2013).
\newblock Overview of the {TAC} 2013 {Knowledge} {Base} {Population}
  {Evaluation}: {English} {Slot} {Filling} and {Temporal} {Slot} {Filling}.
\newblock In {\em Proceedings of the {Text} {Analysis} {Conference}}.

\bibitem[\protect\citename{Tjong Kim~Sang and De~Meulder}2003]{tjong2003}
Tjong Kim~Sang, E.~F. and De~Meulder, F.
\newblock (2003).
\newblock Introduction to the conll-2003 shared task: Language-independent
  named entity recognition.
\newblock In Walter Daelemans et~al., editors, {\em Proceedings of the Seventh
  Conference on Natural Language Learning at HLT-NAACL 2003}, pages 142--147.

\bibitem[\protect\citename{Xu \bgroup et al.\egroup }2007]{xu2007b}
Xu, F., Uszkoreit, H., and Li, H.
\newblock (2007).
\newblock A {Seed-driven} {Bottom-up} {Machine} {Learning} {Framework} for
  {Extracting} {Relations} of {Various} {Complexity}.
\newblock In {\em Proc. of {ACL}}, pages 584--591.

\bibitem[\protect\citename{Yaghoobzadeh and Schütze}2017]{yaghoobzadeh2017}
Yaghoobzadeh, Y. and Schütze, H.
\newblock (2017).
\newblock Multi-level {Representations} for {Fine}-{Grained} {Typing} of
  {Knowledge} {Base} {Entities}.
\newblock In {\em Proceedings of the 15th {Conference} of the {European}
  {Chapter} of the {Association} for {Computational} {Linguistics}: {Volume} 1,
  {Long} {Papers}}, pages 578--589, Valencia, Spain, April. Association for
  Computational Linguistics.

\bibitem[\protect\citename{Zhang \bgroup et al.\egroup }2017]{zhang2017}
Zhang, Y., Zhong, V., Chen, D., Angeli, G., and Manning, C.~D.
\newblock (2017).
\newblock Position-aware attention and supervised data improve slot filling.
\newblock In {\em Proceedings of the 2017 Conference on Empirical Methods in
  Natural Language Processing}, pages 35--45, Copenhagen, Denmark, September.
  Association for Computational Linguistics.

\end{thebibliography}


\end{document}